\pdfoutput=1
\documentclass[11pt]{article}

\usepackage{acl}

\usepackage{times}
\usepackage{latexsym}

\usepackage[T1]{fontenc}
\usepackage[utf8]{inputenc}

\usepackage{microtype}

\usepackage{bbm}
\usepackage{adjustbox}
\usepackage{caption}
\usepackage{subcaption}
\usepackage{verbatimbox}
\usepackage{color, colortbl}
\usepackage{stfloats}
\usepackage{graphicx}
\usepackage{amsmath}

\usepackage{arydshln}
\usepackage{nicefrac}
\usepackage{multirow}
\usepackage{multicol}
\usepackage{amsfonts}
\usepackage{algorithm, algorithmic}
\usepackage{amssymb}
\usepackage{tcolorbox}

\usepackage{hyperref}       % hyperlinks
\usepackage{url}            % simple URL typesetting
\usepackage{booktabs}       % professional-quality tables
\usepackage{amsfonts}       % blackboard math symbols
\usepackage{nicefrac}       % compact symbols for 1/2, etc.
\usepackage{microtype}      % microtypography
\usepackage{xcolor}         % colors

\title{Laziness Is a Virtue When It Comes to \\ Compositionality in Neural Semantic Parsing}

\author{Maxwell Crouse, Pavan Kapanipathi, Subhajit Chaudhury, \\ 
        {\bf Tahira Naseem, Ramon Astudillo, Achille Fokoue, Tim Klinger } \\ 
        \{ maxwell.crouse, subhajit, ramon.astudillo \}@ibm.com \\
        \{ kapanipa, tnaseem, achille, tklinger \}@us.ibm.com
        \\ IBM Research}

\begin{document}
\maketitle
\begin{abstract}
Nearly all general-purpose neural semantic parsers generate logical forms in a strictly top-down autoregressive fashion. Though such systems have achieved impressive results across a variety of datasets and domains, recent works have called into question whether they are ultimately limited in their ability to compositionally generalize. In this work, we approach semantic parsing from, quite literally, the opposite direction; that is, we introduce a neural semantic parsing generation method that constructs logical forms from the bottom up, beginning from the logical form's leaves. The system we introduce is lazy in that it incrementally builds up a set of potential semantic parses, but only expands and processes the most promising candidate parses at each generation step. Such a parsimonious expansion scheme allows the system to maintain an arbitrarily large set of parse hypotheses that are never realized and thus incur minimal computational overhead. We evaluate our approach on compositional generalization; specifically, on the challenging CFQ dataset and three Text-to-SQL datasets where we show that our novel, bottom-up semantic parsing technique outperforms general-purpose semantic parsers while also being competitive with comparable neural parsers that have been designed for each task.
\end{abstract}

\section{Introduction}
\label{introduction}
% In this work, we consider the task of semantic parsing, i.e., the task of parsing a natural language utterance into a structured logical form \cite{zelle1996learning}. The earliest learning-based approaches to semantic parsing would induce probabilistic grammars that defined how words and phrases could be combined into formal meaning representations \cite{zettlemoyer2005learning,Kate2005LearningTT,kwiatkowski2011lexical}. While effective in closed domains with limited vocabularies, such approaches were often challenging to train and difficult to scale to settings involving greater breadth, e.g., WebQuestions \cite{berant2013semantic}. With the advent of deep learning and its application to natural language processing \cite{mikolov2013distributed,sutskever2014sequence}, the semantic parsing community shifted towards neural-based methods for their ability to flexibly handle language and the ease with which they could be trained.

Compositionality is inherent to natural language, with the meaning of complex text or speech understood through the composition of constituent words and phrases \cite{montague1973proper}. For instance, having observed the usage of phrases like "edited by" and "directed by" in isolation, a human would be able to easily understand the question "Was Toy Story edited by and directed by John Lasseter?". The ability to take individual components and combine them together in novel ways is known as \textit{compositional generalization}, and is a key feature of human intelligence that has been shown to play a significant role in why humans are so efficient at learning \cite{lake2017building}.

Compositional generalization has been identified as a major point of weakness in neural methods for semantic parsing \cite{lake2018generalization,higgins2018scan}. Accordingly, this deficiency has been taken up as a challenge by the machine learning community, leading to a slew of methods \cite{liu2021learning,herzig2021unlocking,gai2021grounded} and datasets \cite{keysers2019measuring,kim2020cogs} targeted towards compositional generalization. However, while there has been progress in determining factors that allow systems to compositionally generalize \cite{furrer2020compositional,oren2020improving}, there yet remains gaps in our understanding as to why these neural models fail.

\begin{figure*}[t]
\begin{subfigure}{\textwidth}
\centering
\includegraphics[width=0.85\textwidth]{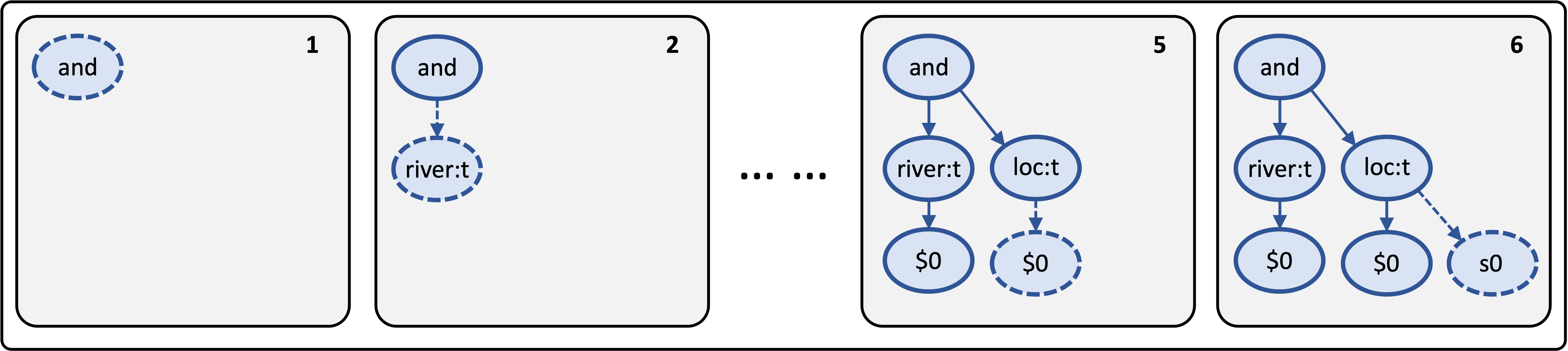}
\caption{Top-down decoding (nodes drawn with dotted lines are those being generated)}
\label{fig:td_gen}
\end{subfigure}
\vskip 0.05in
\begin{subfigure}{\textwidth}
\centering
\includegraphics[width=0.85\textwidth]{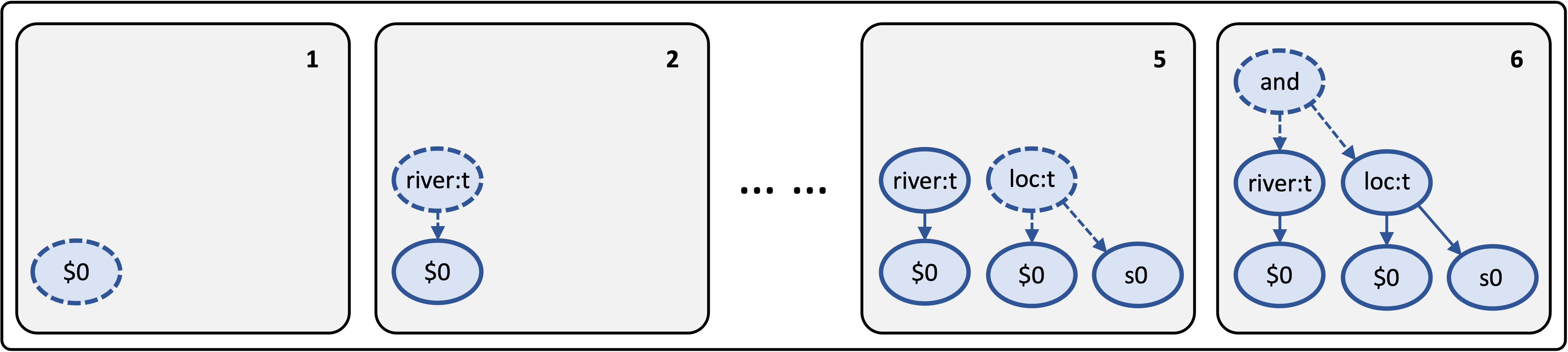}
\caption{Bottom-up decoding (nodes drawn with dotted lines are those being generated)}
\label{fig:bu_gen}
\end{subfigure}
\caption{Top-down versus bottom-up decoding strategies for the Geoquery question "What rivers are in s0?"}
\label{fig:td_bu_diff}
\end{figure*}

%In this paper, we posit that the failure of traditional neural semantic parsers to compositionally generalize is in part due to their entirely autoregressive decoding scheme, wherein the production of each new token is conditioned on all previously generated tokens. In particular, we hypothesize that top-down autoregressive decoding is disadvantageous when generating logical forms that are composed of independent subexpressions, i.e., when there are multiple subexpressions whose meanings are \textit{not} conditional on one another. 
%Here we propose an alternative method for decoding that is designed to address this notion of independence. %The decoder in our model builds up a \textit{set} of possible logical forms, rather than just a single logical form (as with top-down parsers). At each decoding step, logical forms from the set are selected to be joined together or extended with a new symbol (e.g., two entities joined together with a predicate), which allows our method to condition on only the logical expressions involved in that particular generation step. This decoding strategy would ordinarily  \cite{robinson1965machine}
%Specifically, we introduce a new neural decoding architecture that generates logical forms in a bottom-up, semi-autoregressive fashion.

In this paper, we posit that the failure of traditional neural semantic parsers to compositionally generalize is in part due to how they build logical forms. Most commonly, neural semantic parsers treat parsing as an encoder-decoder problem, where the decoder generates logical forms from the top down in an autoregressive fashion \cite{dong-lapata-2016-language,xiao-etal-2016-sequence,AlvarezMelis2017TreestructuredDW,krishnamurthy-etal-2017-neural,dong-lapata-2018-coarse}. That is, these systems output a linearization of the target logical form's abstract syntax tree, beginning from the root of the tree, where the generation of each token is conditioned on both the input text as well as the entire sequence of previously generated tokens.

Such an entirely autoregressive decoding scheme, wherein the production of each new token is conditioned on all previously generated tokens, could result in models that assume invalid dependencies to exist between tokens and thus overfit to their training data (as observed in \cite{qiu-etal-2022-evaluating,bogin-etal-2022-unobserved}). We hypothesize that this overfitting problem would lessen the ability of these models to generalize to unseen compositions.% This overfitting is most problematic in domains where logical forms contain novel combinations of tokens and subexpressions unseen during training. 

%\achille{Is the core problem the presence of unseen sub-expressions or the fact that the traditional approach tends to force invalid dependencies on completely unrelated/independent sub-expressions? My understanding is that the core problem addressed by the bottom-up appraoch is the latter (forced dependencies). Or maybe it is both. Anyway, we need to clarify this as it is one of the core novel aspects of the new approach.}

%\maxwell{The core problem is the invalid dependencies part, I'll clarify this. The unseen combinations of subexpressions then becomes an issue when there are these unnecessary dependencies, but the core issue is the invalid dependencies.}

Here we introduce an alternative decoding approach that eschews the top-down generation paradigm, and instead uses a bottom-up mechanism that builds upwards by combining entities and subexpressions together to form larger subexpressions (Figure \ref{fig:td_bu_diff} demonstrates this distinction). Unlike top-down approaches, our decoder generates logical forms by conditioning on only the relevant subexpressions from the overall graph, hence improving compositional generalizability. 

\textbf{Contributions}: (a) We introduce a novel, bottom-up semantic parsing decoder that achieves strong results, specifically with respect to compositional generalization.
(b) We evaluate our approach on CFQ \cite{keysers2019measuring}, three Text-to-SQL \cite{finegan2018improving} datasets, and Geoquery \cite{zelle1996learning}, and find that it outperforms general-purpose semantic parsers while also being competitive with comparable neural parsers that have been designed for each task. (c) We demonstrate the flexibility of our architecture by testing our approach with multiple encoders, showing that our architecture almost always leads to a significant performance improvement. (d) We show how the bottom-up paradigm can result in a combinatorial explosion in the number of subexpressions created at each decoding step and propose a lazy evaluation scheme that mitigates this by selectively expanding the logical form in a way that minimizes computational overhead.

%Rather than proactively generating all possible subexpressions at each decoding step, it selectively focuses on a single subexpression at a time, generating new additions to . It  expanding from . encoding all possible subexpressions using expensive neural operations, our approach selectively builds logical forms. This lazy scheme allows our approach , separating it from other bottom-up semantic parsing approaches \cite{rubin2021smbop,odena2020bustle,Herzig2021SpanbasedSP} that  beginning from the entities. At each step in decoding, logical forms (or sub-expressions -- see Figure 1) from this set are selected to be extended with a new symbol. 

%\textbf{Contributions:} In this work, . We show that one can simply substitute our decoder into an encoder-decoder semantic parser and a 

\section{Representing Logical Forms}
\label{sec:logical_form}

Our parser is intended to be task-agnostic, with as few as possible assumptions made regarding the particular formalism (e.g., SQL, SPARQL, etc.) a logical form may instantiate. We assume only that we have access to a vocabulary $V$ that defines the set of all symbols a logical form could be constructed from (with no restrictions imposed as to the label or arity of any particular symbol) and that the logical form is given as an s-expression.

S-expressions are a representation formalism that expresses trees as nested lists (e.g., \texttt{(count \$0 (river:t \$0))} as a query for the question "How many rivers are there?"). Their use dates back to the earliest days of artificial intelligence research \cite{mccarthy1983recursive} as the main unit of processing of the LISP programming language \cite{mccarthy1978history}. In this work, their main purpose is to simplify the structure of logical forms into binary trees. Importantly, the transformation of logical forms into s-expressions requires no knowledge beyond the syntax of the target formalism.

\begin{figure}[t]
\begin{subfigure}{\columnwidth}
\centering
\includegraphics[width=\columnwidth]{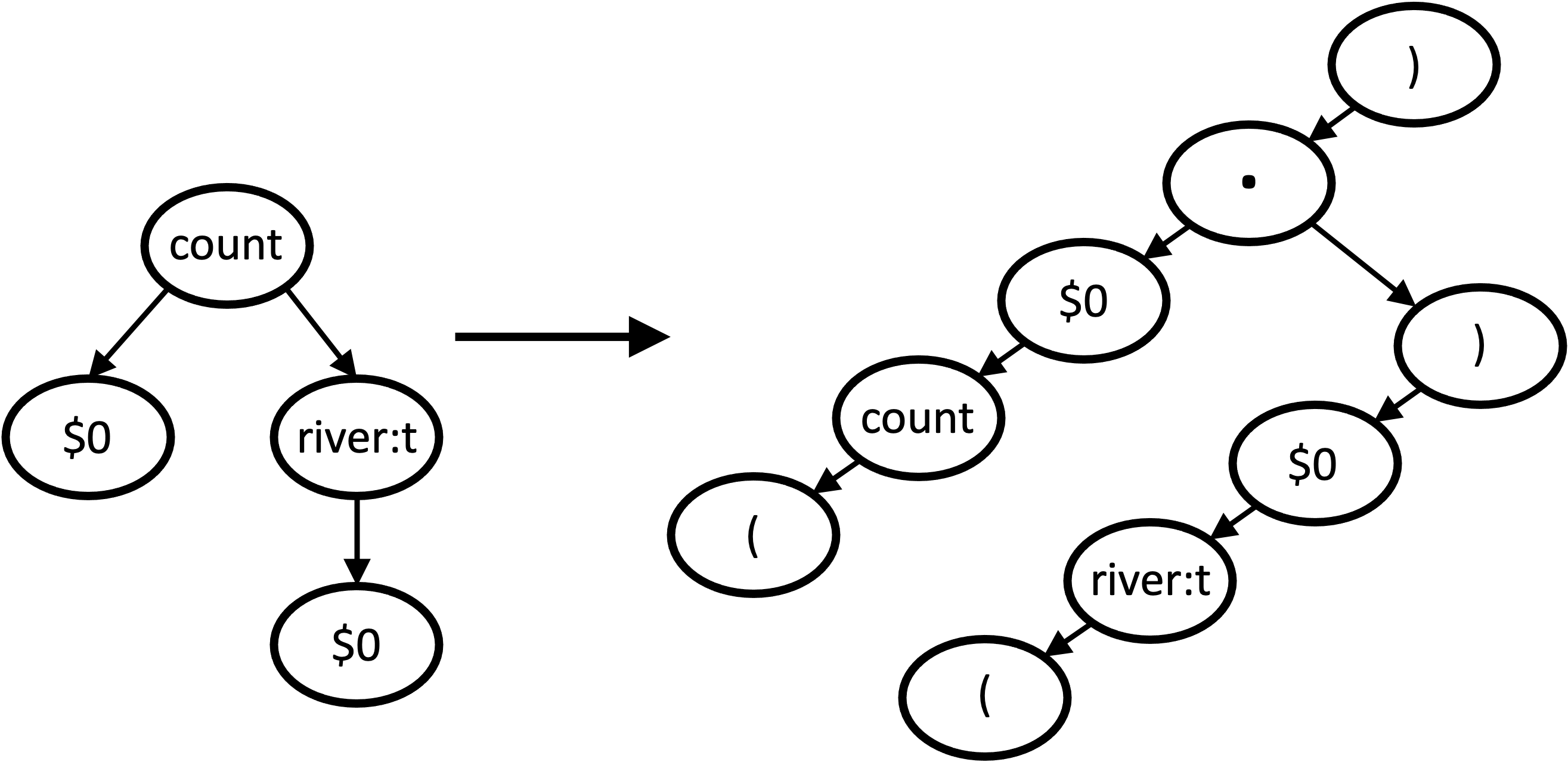}
\caption{Internal conversion of the logical form \texttt{(count \$0 (river:t \$0))} into a tree-based s-expression for the question "How many rivers are there?"}
\label{fig:sexpr}
\end{subfigure}
\vskip 0.1in
\begin{subfigure}{\columnwidth}
\centering
\includegraphics[width=\columnwidth]{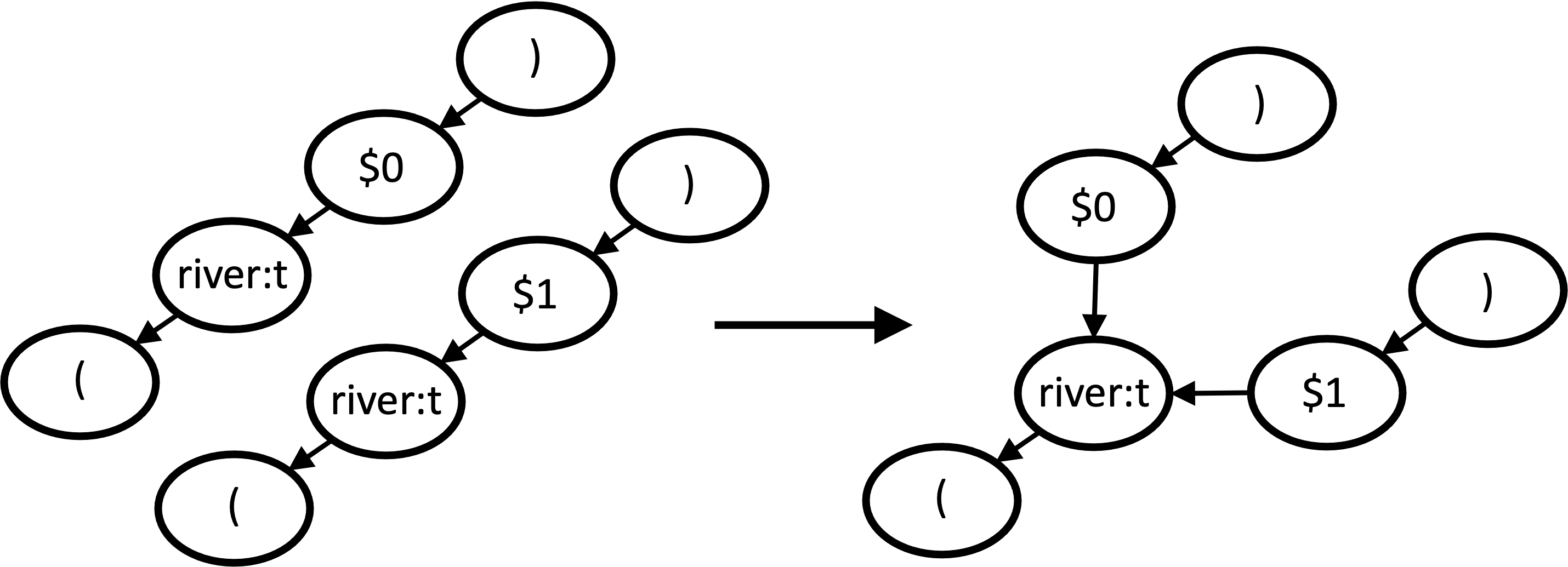}
\caption{DAG representation of an s-expression where identical subtrees have been mapped to the same nodes}
\label{fig:dag_sexpr}
\end{subfigure}
\caption{Preprocessing graph transformations}
\end{figure}

Traditionally, logical forms are generated as trees; however, we instead represent them as directed acyclic graphs (DAGs). This is a logically equivalent representation that is created by collapsing all identical subtrees into a single subgraph that maintains connections to each of the original parent nodes. Figure \ref{fig:sexpr} provides an example of how an s-expression would initially be converted into a tree representation, while Figure \ref{fig:dag_sexpr} shows how two overlapping trees would be merged into a single DAG form. Within a graph all nodes will either have 1) one argument and a label drawn from the vocabulary of symbols \textit{V} or 2) two arguments and connect an incomplete list (its left argument) with a complete list (its right argument). The nodes with two arguments will always use a special pointer symbol ``$\boldsymbol{\cdot}$'' as their label.

As the final preprocessing step, each logical form is wrapped in a special \texttt{root} s-expression. For instance, the logical form \texttt{(count \$0 (river:t \$0))} would become \texttt{(root (count \$0 (river:t \$0)))}. This step provides a signal to the model to stop generating new extensions to the logical form. During decoding, only those s-expressions that begin with a \texttt{root} token may be returned to the user.

\section{Model}
\label{model}

Given a question $Q$, our method is tasked with producing a graph representation $G$ of a logical form using symbols drawn from a vocabulary $V$. At the highest level, our approach follows the traditional encoder-decoder paradigm. It first parses the input with a neural encoder (e.g., LSTM \cite{hochreiter1997long}) to produce real-valued vector representations for each word in $Q$. Those representations are then passed to a neural decoder, which iteratively takes decoding actions to generate $G$. Our approach is agnostic to the choice of encoder, and thus in the following sections we will describe only the decoding process. 

An important desideratum of this work was architectural simplicity. Thus, when using a pretrained large language model as the base to our approach, (e.g., T5 \cite{raffel2019exploring}), the model was left largely as is. None of the modifications described in the following sections involve changes to the internal parameters or architecture of the neural model itself (i.e., all additions are kept to the output layer). Consequently, when our approach is instantiated with a pretrained language model, that model is applied entirety off-the-shelf.

\subsection{Decoding Actions}

\begin{figure}[t]
\centering
\includegraphics[width=0.5\columnwidth]{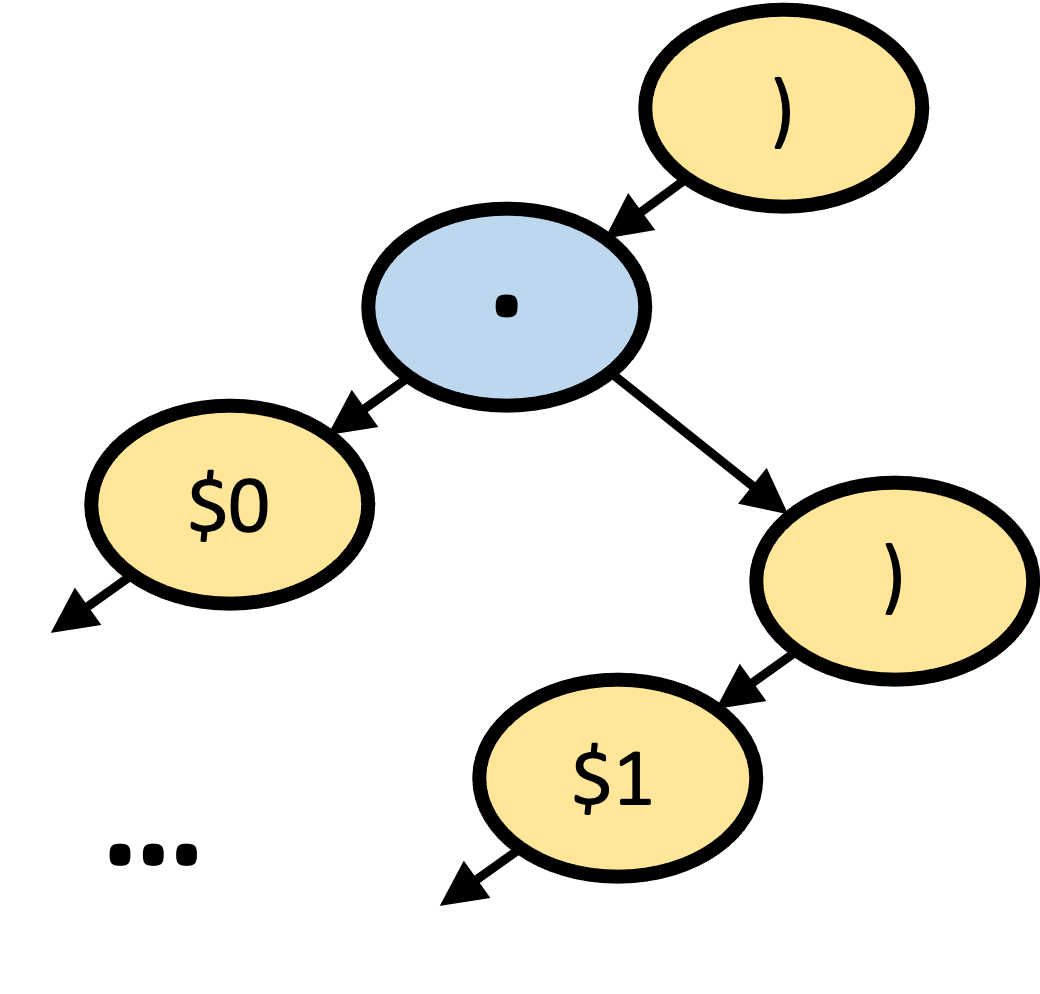}
\caption{Nodes in $G$ originating from generation actions (yellow) and from pointer actions (blue).}
\label{fig:gen_ptr_nodes}
\end{figure}

At each step of decoding, our system executes either a \textit{generation action} or a \textit{pointer action}. Generation actions take a single node from $G$ and add it as the argument to a new node with a label drawn from $V$. Pointer actions instead take two nodes from $G$ and add them both as arguments to a new node with a special pointer symbol ``$\boldsymbol{\cdot}$'' as its label. The division of decoding into generation and pointer actions is adapted from \cite{zhou2021amr,zhou2021structure}, which proposes this scheme to drive a transition-based AMR parser. Key to note is that both action types will result in a new node being generated that has as arguments either one or two already-existing nodes from $G$, i.e., a bottom-up generation process. Figure \ref{fig:gen_ptr_nodes} illustrates the outcomes of both action types.

At the start of decoding our approach initializes two objects, $G$ as the empty graph and a set of candidate actions $A~=~\{ \left< \ \texttt{(} \ , \emptyset, 1 \right> \}$. The first action of $A$ will always be to generate the left parenthesis, which is the start to every s-expression in $G$. Each decoding step begins by extending $G$ with actions drawn from $A$ and ends with a set of new candidate actions being added to $A$.

The key to our approach lies in its \textit{lazy} evaluation scheme, wherein actions that create nodes are generated at each decoding step, rather than the nodes themselves. This allows our model to strongly restrict how $G$ expands at each decoding step, with our approach being able to build up a large set of unexecuted actions representing candidate logical forms that are never processed with resource-intensive neural components and thus incur very little computational overhead.

\subsection{Candidate Selection}

\begin{figure}[t]
\begin{center}
\includegraphics[width=\columnwidth]{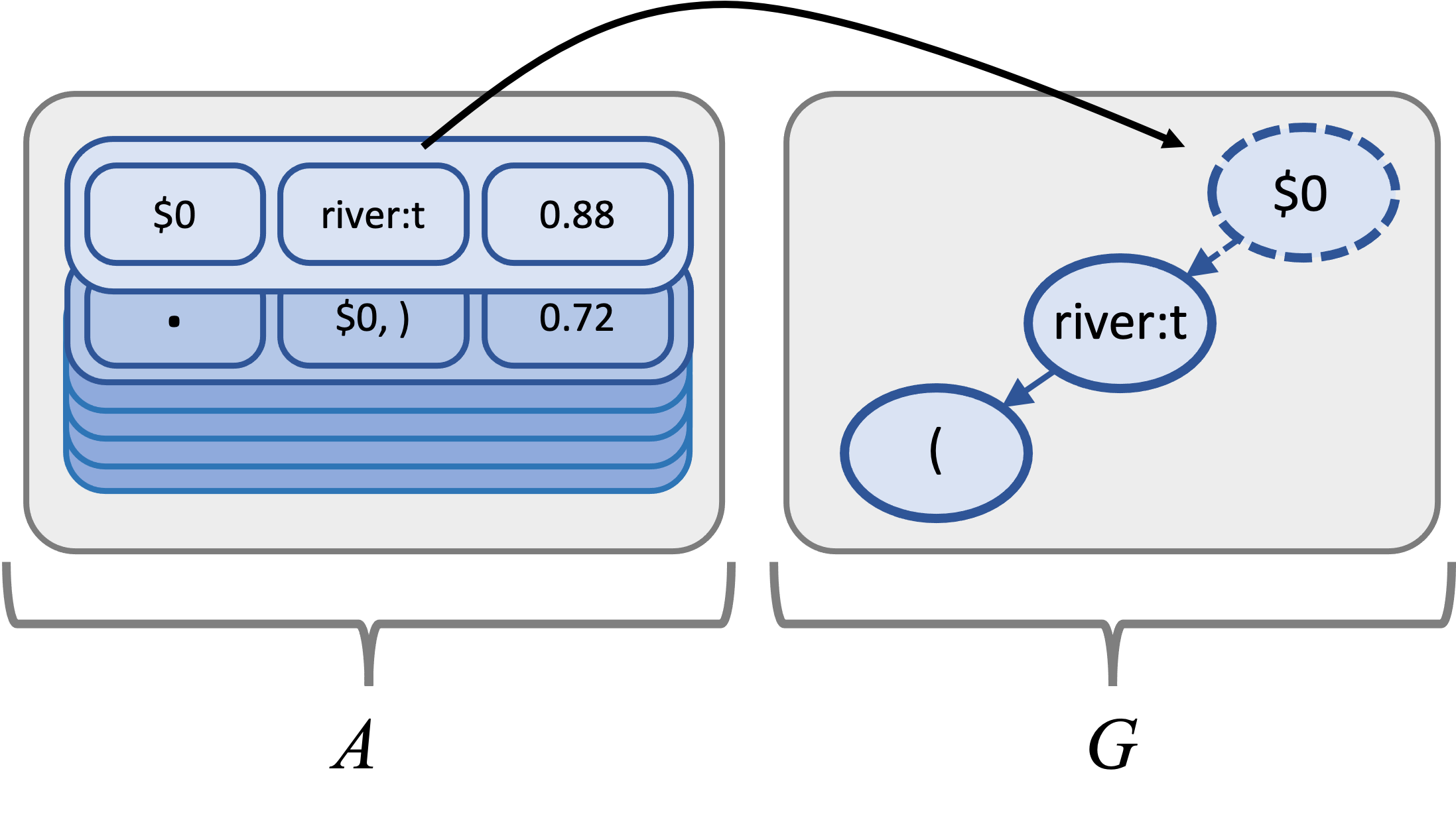}
\caption{A generation action from \textit{A} that, when selected and executed, adds a node with label \texttt{\$0} to \textit{G}}
\label{fig:add_node}
\end{center}
\end{figure}

Each element $a \in A$ is a tuple $a = \left<v, \mathcal{A}, p_a\right>$ consisting of: 1) a symbol $v$ drawn from our vocabulary $V$, 2) an ordered list of arguments $\mathcal{A}$, where each member of $\mathcal{A}$ is a node in $G$, and 3) a probability $p_a$ reflective of the model's confidence that this candidate should be part of the final returned output. Adding a candidate $a = \left<v, \mathcal{A}, p_a\right>$ to $G$ simply involves creating a new node labeled $v$ within $G$ that has directed edges connecting the node to each argument in $\mathcal{A}$, as Figure \ref{fig:add_node} demonstrates.

Our model is equipped with a selection function that will select and remove members of $A$ to add to $G$. In the experiments of this paper, our selection function chooses all elements of $A$ with a probability above a certain threshold $\kappa$ (e.g., $\kappa = 0.5$) or, if no such options exist, chooses the single highest probability option from $A$.

\subsection{Graph Encoding}

\label{sec:decoder}

For our neural model to make predictions, it must first convert the newly generated nodes into real-valued vector representations. The decoder of our model largely follows the design of T5 \cite{raffel2019exploring}, i.e., a transformer \cite{vaswani2017attention} using relative positional encodings \cite{shaw2018self}, but with modifications that allow it to process graph structure. As most equations are the same as those detailed in the original T5 work, we will only describe the differences between our model and theirs.

Throughout the remainder of this section, we will define a frontier $F$ to refer to the set of nodes generated by actions in $A$ that were selected during the current round of decoding. Each element of $F$ is first assigned a vector embedding according to its label. The embedding is then passed through the decoder to produce real-valued vector representations for each node. In order to capture the structure of $G$ with the transformer, our model makes use of two types of information within the attention modules of each layer of the decoder.

\begin{figure}[t]
\begin{center}
\includegraphics[width=\columnwidth]{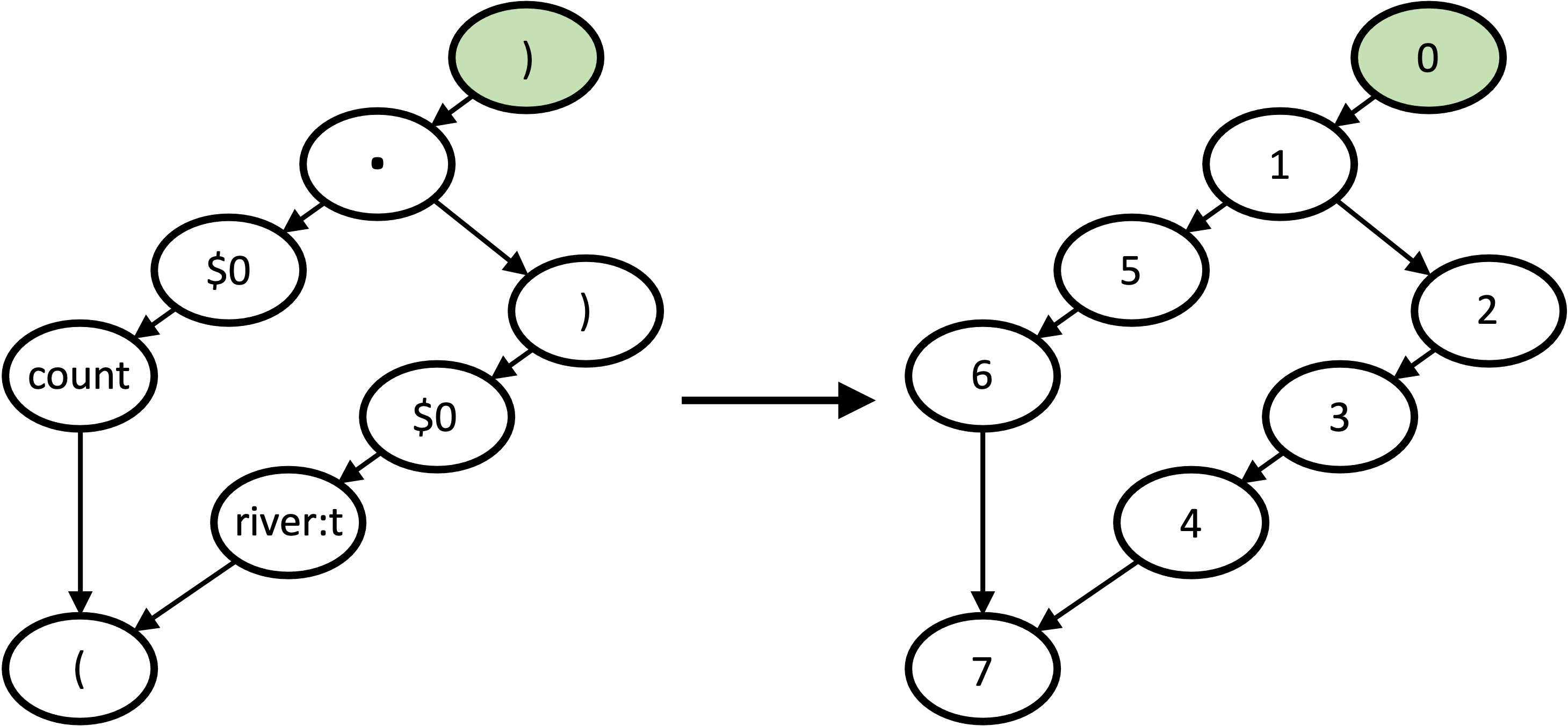}
\caption{Reversed post-order traversal-based position assignment relative to a newly generated node (colored in green). The number assigned to each node is the index of the position bias (e.g., 0 would be assigned to bias $b_0$, 1 would be assigned to bias $b_1$, etc.)}
\label{fig:pos_bias}
\end{center}
\end{figure}

First, within the self-attention module, a node $n \in G$ may only attend to its set of descendants, i.e., only the nodes contained within the subgraph of $G$ rooted at $n$. This has the effect that the resultant encoding of a node is a function of only its descendants and the original text. Second, for a pair of nodes $n_i$ and $n_j$ for which an attention score is being computed (e.g., $n_i$ is the parent of $n_j$), the positional encoding bias $b_{ij}$ used to represent the offset between $n_i$ and $n_j$ is assigned according to a reverse post-order traversal of the descendants of node $n_i$. That is, $b_{ij}$ is selected according to where $n_j$ is positioned in a post-order traversal of the subgraph of $G$ rooted at $n_i$. Figure \ref{fig:pos_bias} provides an example of this ordering scheme for a particular node. 

This assignment strategy effectively linearizes the graph rooted at $n_i$ and has the added property that each descendant of $n_i$ will be assigned a unique position bias with respect to $n_i$. This is also a somewhat analogous encoding as to what is done for standard seq-to-seq models. To see the similarity, consider a "linear" graph (e.g., a sequence); the encoding for each node is exactly that given by a distance-based positional encoding \cite{shaw2018self} (e.g., used by T5).

\subsection{Action Generation}
\label{sec:generation}
\begin{figure}[t]
\begin{center}
\includegraphics[width=\columnwidth]{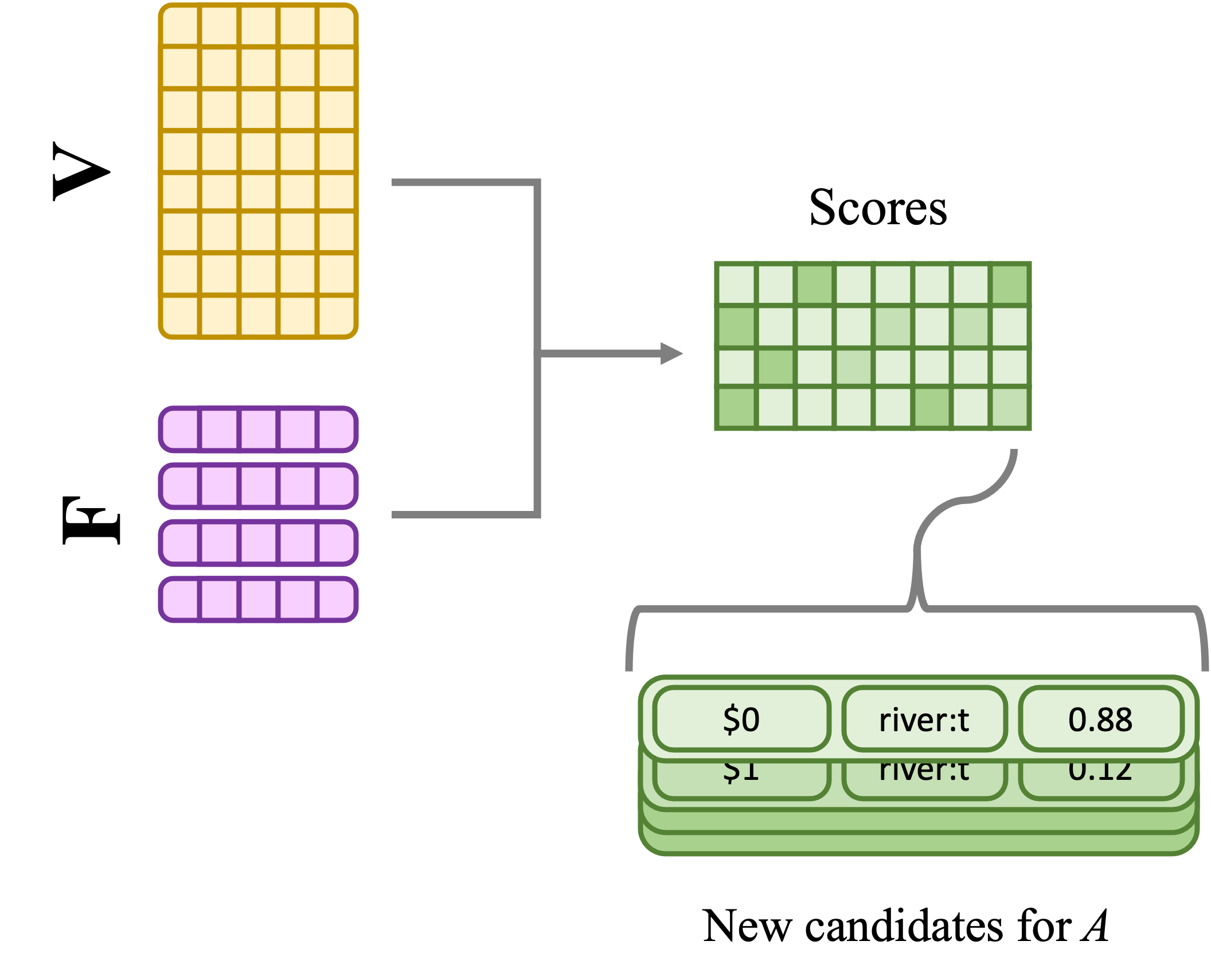}
\caption{The embeddings of nodes in the frontier \textit{F} and the embeddings of symbols in the vocabulary \textit{V} being used to construct new generation actions}
\label{fig:gen_action}
\end{center}
\end{figure}

Once the decoder has processed each node, our approach will have a set of node embeddings $\{ h_1, h_2, \ldots h_{|F|} \}$. To produce the set of decoding actions, our model executes two operations (one for each action type). The first operation proposes a set of generation actions (a brief depiction of this is given in Figure \ref{fig:gen_action}). For a node $n_i$ and symbol $v$ with embedded representations $h_i$ and $h_v$, respectively, this involves computing
\begin{alignat*}{2}
&p_{i}^{(v)} &&= \sigma \big( h_i^{\top} h_{v} + b \big)
\end{alignat*}
where $b$ is a bias term and $\sigma$ is the sigmoid function. The value $p_{i}^{(v)}$ can be interpreted as the \textit{independent} probability (independent of all other action probabilities) that $G$ should contain a node with label $v \in V$ that has $n_i$ as an argument. For each $v \in V$, $A$ is extended as $A = A \cup \{ \big<v, \left< n_i \right>, p_{i}^{(v)} \cdot p_{n_i} \big> \}$, where $p_{n_i}$ refers to the probability of the \textit{original} action that generated $n_i$.

The second operation proposes a set of pointer actions using an attention mechanism. For a pair of nodes $\big<n_i, n_j\big>$ with embedded representations $h_i$ and $h_j$, respectively, this is computed as
\begin{alignat*}{2}
& q_{i} &&= W^{(q)} h_i,\;\;\;\;k_{j} = W^{(k)} h_{j} \\
&p_{i}^{(j)} &&= \sigma \big( \dfrac{q^{\top}_{i} k_j}{\sqrt{d}} + b \big)
\end{alignat*}
where $W^{(q)}$, $W^{(k)}$ are learned matrices, $b$ is a bias term, $d$ is the dimensionality of the embeddings, and $\sigma$ is the sigmoid function. Like before, $p_{i}^{(j)}$ can be thought of as the \textit{independent} probability that $G$ will contain a node that has $n_i$ as its left argument and $n_j$ as its right argument. For each pair $\left< n_i, n_j \right> \in (F \times G) \cup (G \times F)$ (where $F \cup G$ is a slight abuse of notation to write the union between $F$ and the set of nodes within $G$), we update the set of actions $A = A \cup \{ \big< \ \boldsymbol{\cdot} \ , \left< n_i, n_j \right>, p^{(j)}_{i} \cdot p_{n_i} \cdot p_{n_j} \big> \}$, where $p_{n_i}$ and $p_{n_j}$ refer to the probabilities of the original actions that generated $n_i$ and $n_j$.

It is important to note the combinatorial nature of generating new actions. At each round of decoding, the number of actions could grow by $O(|F|\cdot|V|+|F|\cdot|G|)$. While it is certainly possible to restrict the number of actions added to \textit{A} (e.g., by adding only the top-$k$ actions), the set still grows extremely quickly. 

A key feature to our decoding scheme is that it does not actually build and embed the graphs for a particular action until that action is executed. Because each action maintains the probability of its source actions, our model is effectively exploring via Dijkstra's algorithm, where the frontier \textit{F} includes all actions yet to be executed. This contrasts with previous works, e.g., \cite{rubin2021smbop}, which are greedier in that they keep a fixed beam with new nodes that are added and embedded on each round and all nodes outside this beam being discarded (our approach does not need to discard these nodes since they have no significant cost until they are executed).

%A key feature of our approach is that the actions within \textit{A} need not be evaluated. If one were to eagerly evaluate the actions and add the resulting nodes to $G$ (as is done in other bottom-up parsing works, e.g., \cite{rubin2021smbop}) there would be a significant  computational burden involved in embedding each new node.

\subsection{Terminating Decoding}

Decoding terminates when an s-expression beginning with the \texttt{root} token is generated (refer to the end of Section \ref{sec:logical_form} for an example). In order to ensure the highest probability s-expression is returned, decoding only terminates if the final action's probability is also the highest amongst all yet-unselected actions in $A$. Upon generating the \texttt{root} s-expression, only the subgraph of $G$ rooted at the newly generated node is returned. Thus, though the the size of $G$ may grow to be quite large, not every element of $G$ is returned.

\subsection{Training}

Training our model is as efficient as other standard transformer-based models, i.e., a single parallel operation that processes the entire graph and computes all action probabilities simultaneously. Additionally, the memory footprint of our model is practically equivalent to the underlying transformer used to initialize its parameters (e.g., T5).

The loss function used to train our model is a cross-entropy-based loss. Recall that both generation and pointer action probabilities are the output of a sigmoid function. Letting $Q$ be the input question, $\mathcal{P}$ be the set of positive actions (i.e., actions needed to generate the gold-annotation logical form), and $\mathcal{N}$ be the set of negative actions (i.e., actions not needed to generate the gold-annotation logical form), the objective is written as
\begin{alignat*}{2}
&\mathcal{L} &&= \sum_{n \in \mathcal{P}} \log{p_\theta(n | Q)} + \sum_{n \in \mathcal{N}} \log{1 - p_\theta(n | Q)}
\end{alignat*}
where the conditional probabilities $p_\theta(n | Q)$ are the sigmoid-constrained outputs of generation (see Section \ref{sec:generation}) and $\theta$ is the set of model parameters. 

While $\mathcal{P}$ is fixed and finite, the set $\mathcal{N}$ is clearly unbounded. In this work, we construct $\mathcal{N}$ from a small set of both generation and pointer actions. For each node $n \in G$, we add negative generation actions for each symbol $v \in V$ that is \textit{not} the label of a parent of $n$. For negative pointer actions, we add actions for each pair of nodes in $G$ that do not have a common parent.

\section{Experiments}
\label{experiments}

\begin{figure}[t]
\begin{subfigure}{\columnwidth}
\begin{center}
\begin{tcolorbox}[width=0.95\columnwidth,colback=white!90!black]
{\texttt{(SELECT (count *) (WHERE \\ \hspace*{1pt} (directed\_by M0 M1) \\ \hspace*{1pt} (directed\_by M0 M2)\\ \hspace*{1pt} (edited\_by M0 M1)\\ \hspace*{1pt} (edited\_by M0 M2)\\ \hspace*{1pt} (prod\_companies M0 M1)\\ \hspace*{1pt} (prod\_companies M0 M2)))
}}
\end{tcolorbox}
\end{center}
\end{subfigure}

\caption{S-expression for the CFQ question "Was M0 produced, directed, and edited by M1 and M2"}
\label{fig:cfq_repr}
\end{figure}

In our experiments, we aimed to answer the following questions: 1) is bottom-up a more effective decoding paradigm than top-down with respect to compositional generalization and 2) how well does our approach generalize to different domains and formalisms. To answer these questions, we evaluated our approach on several datasets: 1) CFQ \cite{keysers2019measuring}, 2) the SQL versions of Geoquery \cite{zelle1996learning}, ATIS \cite{dahl1994expanding}, and Scholar \cite{iyer2017learning}, each of which were curated by \cite{finegan2018improving}, and 3) the version of Geoquery \cite{zelle1996learning} used by \cite{dong-lapata-2016-language,kwiatkowski2011lexical} which maps questions to a variant of typed lambda calculus \cite{Carpenter1997TypeLogicalS}. To save space, we provide all hyperparameters used for our experiments in the appendix.

Our main experiments investigating compositional generalization were performed on the CFQ and Text-to-SQL datasets. In CFQ, the distribution of train and test data in each split is designed to exhibit maximum compound divergence (MCD) \cite{keysers2019measuring}. In the MCD setting, data is comprised of a fixed set of atoms, i.e., the primitive elements to a question. But while each atom is observed in training, the compositions of atoms between train and test sets will vary significantly (an example of a CFQ question is provided in Figure \ref{fig:cfq_repr}). In the Text-to-SQL datasets, the train and test sets are constructed by using separate template splits \cite{finegan2018improving,herzig2021unlocking}, where the train and test sets have questions involving different SQL query templates.

%Figure \ref{fig:cfq_repr} provides an example, which shows the query associated with a question from the CFQ dataset \cite{keysers2019measuring}. In the example, the atoms are the individual expressions within the \texttt{WHERE} statement. For a system to compositionally generalize on this dataset, it must learn representations of the individual atoms for both the text (e.g., "directed" and "edited by") as well as the logical forms (e.g., \texttt{directed\_by} and \texttt{edited\_by}) that can be applied in contexts dissimilar to those they were trained on.

%  For the text-to-SQL datasets,
% we use template splits (Finegan-Dollak et al., 2018),
% which ensure that the train and test set contain distinct SQL query “templates” (constructed by replacing values in the SQL queries with anonymized
% placeholders).

\begin{table*}[t]
\begin{center}
\begin{small}
\begin{tabular}{lcccc}
\toprule
Method & MCD1 & MCD2 & MCD3 & MCD Avg. \\
\midrule
Dynamic Least-to-Most Prompting \cite{drozdov2022compositional}& 94.3 & 95.3 & 95.5 & 95.0  \\
LEAR \cite{liu2021learning}& 91.7 & 89.2 & 91.7 & 90.9  \\
LIR+RIR (T5-3B) \cite{herzig2021unlocking} & 88.4 & 85.3 & 77.9 & 83.8  \\
Grounded Graph Decoding \cite{gai2021grounded} & 98.6 & 67.9 & 77.4 & 81.3 \\
HPD \cite{guo2020hierarchical} & 79.6 & 59.6 & 67.8 & 69.0 \\
LIR+RIR (T5-base) \cite{herzig2021unlocking} & 85.8 & 64.0 & 53.6 & 67.8 \\
T5-11B-mod \cite{furrer2020compositional} & 61.6 & 31.3 & 33.3 & 42.1 \\
LAGR \cite{jambor2021lagr} & 57.9 & 26.0 & 20.9 & 34.9 \\
\midrule
T5-3B \cite{herzig2021unlocking} & 65.0 & 41.0 & 42.6 & 49.5 \\
T5-base \cite{herzig2021unlocking} & 58.5 & 27.0 & 18.4 & 34.6 \\
Edge Transformer \cite{bergen2021systematic} & 47.7 & 13.1 & 13.2 & 24.7 \\
Evolved Transformer \cite{keysers2019measuring} & 42.4 & 9.3 & 10.8 & 20.8 \\
Universal Transformer \cite{keysers2019measuring} & 37.4 & 8.1 & 11.3 & 18.9 \\
Transformer \cite{keysers2019measuring} & 34.9 & 8.2 & 10.6 & 17.8 \\
LSTM \cite{keysers2019measuring} & 28.9 & 5.0 & 10.8 & 14.9 \\
\midrule
LSP (Ours) & & & & \\
\ \ \ \ + T5-base & 88.7 & 57.7 & 43.8 & 63.4 \\
\ \ \ \ + LSTM Encoder & 73.0 & 29.5 & 23.1 & 41.9 \\
\bottomrule
\end{tabular}
\end{small}
\end{center}
\caption{Performance across different splits of the CFQ dataset. Those systems leveraging CFQ-specific algorithms or representations are grouped in the top half of the table, while those systems in the bottom half are domain general}
\label{res:cfq_results}
\end{table*}

\begin{table}[t]
\footnotesize
\begin{center}
\begin{small}
\begin{tabular}{lccc}
\toprule
Method & ATIS & Geoquery & Scholar \\
\midrule
Seq2Seq $\clubsuit$ & 32.0 & 20.0 & 5.0 \\
GECA $\diamondsuit$ & 24.0 & 52.1 & -- \\
Seq2Seq $\spadesuit$ & 28.0 & 48.5 & -- \\
Transformer $\spadesuit$ & 23.0 & 53.9 & -- \\
Seq2Seq+ST $\spadesuit$ & 29.1 & 63.6 & -- \\
Transformer+ST $\spadesuit$ & 28.6 & 61.9 & -- \\
SpanBasedSP $\square$ & -- & 82.2 & -- \\
\midrule
Baseline (T5-base) $\heartsuit$ & 32.9 & 79.7 & 18.1 \\
Baseline (T5-large) $\heartsuit$ & 31.4 & 81.9 & 17.5 \\
Baseline (T5-3B) $\heartsuit$ & 29.7 & 79.7 & 16.2 \\
LIR+RIR (T5-base) $\heartsuit$ & 47.8 & 83.0 & 20.0 \\
LIR+RIR (T5-large) $\heartsuit$ & 43.2 & 79.7 & 22.0 \\
LIR+RIR (T5-3B) $\heartsuit$ & 28.5 & 75.8 & 12.4 \\
\midrule
LSP (Ours) & & \\
\ \ \ \ + T5-base & 38.3 & 81.3 & 25.1 \\
\bottomrule
\end{tabular}
\end{small}
\end{center}
\caption{Performance on Text-to-SQL against: $\clubsuit$\cite{finegan2018improving}, $\diamondsuit$\cite{andreas2020good}, $\spadesuit $\cite{zheng2020compositional}, $\heartsuit$\cite{herzig2021unlocking}, $\square$\cite{Herzig2021SpanbasedSP} }
\label{res:sql_results}
\end{table}

\begin{table}[t]
\footnotesize
\begin{center}
\begin{small}
\begin{tabular}{lc}
\toprule
Method & Acc. \\
\midrule
DCS+L w/ lexicon \cite{liang2013learning} & 91.1 \\
TISP \cite{zhao2015type} & 88.9 \\
KZGS11 \cite{kwiatkowski2011lexical} & 88.6 \\
DCS+L w/o lexicon\cite{liang2013learning} & 87.9 \\
AQA \cite{crouse2021question} & 87.5 \\
$\lambda$-WASP \cite{wong2007learning} & 86.6 \\
ZC07 \cite{zettlemoyer2007online} & 86.1 \\
\midrule
AWP + AE + C2 \cite{jia-liang-2016-data} & 89.3 \\
Graph2Tree \cite{li2020graph} & 88.9 \\
Coarse2Fine \cite{dong-lapata-2018-coarse} & 88.2 \\
Seq2Tree \cite{dong-lapata-2016-language} & 87.1 \\
SpanbasedSP \cite{Herzig2021SpanbasedSP} & 86.4 \\
Graph2Seq \cite{xu2018graph2seq} & 85.7 \\
\midrule
LSP (Ours) & \\
\ \ \ \ + LSTM Encoder & 86.4 \\
\bottomrule
\end{tabular}
\end{small}
\end{center}
\caption{Performance on Geoquery, with neural-based approaches grouped in the bottom half of the table}
\label{res:geo_results}
\end{table}

\subsection{Dataset-Specific Processing}

Task-specific representations and encoding schemes are very common with state-of-the-art approaches for compositional generalization \cite{shaw2021compositional}. Most often, such approaches will transform the target logical form to better reflect some characteristic of the input text. For instance, with CFQ, systems will transform their target logical forms into a query that closely aligns with the list-heavy syntactic structure of CFQ questions \cite{liu2021learning,herzig2021unlocking,jambor2021lagr,furrer2020compositional}.

As our objective was to determine the effectiveness of our bottom-up technique for general-purpose semantic parsing, we avoided any transformation that involved task-specific knowledge about the input text. For the Text-to-SQL and Geoquery datasets, we performed no processing of the inputs or outputs beyond the transformation of input logical forms into s-expressions. For CFQ, we found it necessary to apply two preprocessing steps. First, to mitigate the number of candidate logical forms proposed, only one \texttt{WHERE} clause was allowed to be generated during decoding. Second, we observed that, because we restricted the number of \texttt{WHERE} clauses generated, our method often prematurely ended generation. Thus, we found it useful to add any completed s-expression (i.e., s-expressions ending with the \texttt{)} token) to the \texttt{WHERE} clause of the final logical form.

\subsection{Evaluation}

We evaluated our approach using exact match accuracy, i.e., whether the generated logical form exactly matches the gold logical form annotation. To accommodate for unordered $n$-ary predicates, we reordered the arguments to all such unordered predicates (e.g., \texttt{and}) lexicographically in both the gold and generated logical forms before comparing them. The unordered predicates are the \texttt{WHERE} operator in CFQ as well as the \texttt{and} predicate in the lambda calculus version of Geoquery.

\section{Results and Discussion}
\label{results}

Table \ref{res:cfq_results} shows the results of our approach, referred to as LSP (\textbf{L}azy \textbf{S}emantic \textbf{P}arser), on the CFQ dataset. As can be seen from the table, our approach fares quite well against domain-general semantic parsing approaches and, importantly, significantly outperforms both T5-base and the LSTM.

The large performance increase as compared to both the basic LSTM and T5-base supports our hypothesis that a bottom-up, semi-autoregressive decoding strategy is a better inductive bias for compositional generalization than autoregressive, top-down decoding. However, a better decoding strategy clearly does not provide the complete solution, as evidenced by the gap in performance between the best systems designed specifically for compositional generalization and our approach.

Table \ref{res:sql_results} shows the performance of our approach on the three Text-to-SQL datasets of \cite{finegan2018improving}. While our approach clearly outperforms the baselines, it has mixed results as compared to the work of \cite{herzig2021unlocking}. Additionaly, our approach was outperformed by \cite{Herzig2021SpanbasedSP} on the Geoquery SQL dataset; however we note that their approach required a manually constructed lexical mapping of text to Geoquery terms (they reached 65.9\% accuracy without the mapping).  Still, that our approach outperformed T5-base in all three datasets again supports our hypothesis that bottom-up is more effective for compositional generalization.

\begin{table*}[t]
\footnotesize
\begin{center}
\begin{small}
\begin{tabular}{lcccccc}
\toprule
Method & MCD1 & MCD2 & MCD3 & ATIS & Geoquery & Scholar \\
\midrule
T5-base & 58.5 & 27.0 & 18.4 & 32.9 & 79.7 & 18.1 \\
RIR & 86.3 & 49.1 & 46.8 & 81.3 & 36.3 & 19.4 \\
LIR$_d$ & 48.1 & 40.3 &  35.3 & 44.4 & 83.5 & 20.6 \\
LIR$_d$+RIR & 72.5 & 61.1 & 51.2 & 47.8 & 83.0 & 20.0 \\
LIR$_{ind}$ & 57.6 & 41.4 &  34.7 & 38.3 & 80.8 & 16.5 \\
LIR$_{ind}$+RIR & 85.8 & 64.0 & 53.6 & 41.5 & 81.9 & 16.5 \\
\midrule
LSP (Ours) & & & & & \\
\ \ \ \ + T5-base & 88.7 & 57.7 & 43.8 & 38.3 & 81.3 & 25.1 \\
\bottomrule
\end{tabular}
\end{small}
\end{center}
\caption{Performance as compared to task-specific T5-base models from \cite{herzig2021unlocking}.}
\label{res:herzig_results}
\end{table*}

Table \ref{res:geo_results} shows the performance of our approach on Geoquery. To keep the comparison relatively fair, we used only an LSTM encoder and randomly initialized token embeddings for this experiment. Despite using an entirely different generation scheme, our method is competitive with the other neural-based semantic parsers on this dataset. Notably, there is not a significant drop from Seq2Tree \cite{dong-lapata-2016-language}, which can be considered the antithesis to our method as it decodes trees from the top down. We find these results to be very encouraging, as our approach maintains competitive performance against a varied set of approaches that leverage hand-engineered data augmentation strategies \cite{jia-liang-2016-data}, pretrained word embeddings \cite{li2020graph}, and hand-built lexicons \cite{Herzig2021SpanbasedSP}.

Lastly, in Table \ref{res:herzig_results} we highlight the T5-base models of \cite{herzig2021unlocking}, which used task-specific intermediate representations to simplify the learning problem of semantic parsing for an off-the-shelf large language model. This is a very useful work to compare against, as the number of parameters between our model and theirs is near identical. From the table, we see that our approach roughly matches that of their best performing T5-base model. This is an interesting result, as it could suggest that our architecture is naturally capturing the useful properties of their intermediate representations without as much need for hand-engineering. Importantly, unlike their approach which assumes knowledge of the syntactic structure of questions, ours needs only a rule for identifying expressions (e.g., parentheses delineate expression boundaries) or a grammar for the target (readily available for formalisms like SPARQL or SQL).

\section{Related Work}
\label{related_work}

While the overall parsing approach of \cite{zhou2021amr,zhou2021structure} is quite different from ours (as they propose a transition-based semantic parser), several aspects of our work were inspired from their ideas. For instance, as mentioned in Section \ref{sec:decoder}, they proposed the use of two types of decoding actions to expand the target graph. 

Bottom-up neural semantic parsers are a relatively recent development. Most related to our work is that of \cite{rubin2021smbop}, which introduced a bottom-up system that achieved near state-of-the-art results on the Spider dataset \cite{Yu2018SpiderAL} while being significantly more computationally efficient than its top-down, autoregressive competitors. There are three significant differences between their approach and ours. 
First, they generate trees rather than DAGs and thus assume a many-to-one correspondence between input spans in the question and tokens in the logical form (which does not hold for datasets like CFQ). 
Second, they use an eager evaluation scheme (i.e., nodes are generated at each decoding step and not actions), and thus enforce a beam-size hyperparameter $K$ that must be defined a priori and places a hard limit on the number of subtrees that can be considered at a given time. 
Third, they use a highly customized decoder which limits how tightly integrated their approach can be with off-the-shelf language models. In contrast, when using a pretrained large language model as the base to our approach, we reuse the entirety of the language model as is to instantiate our model.

Another similar work is \cite{Herzig2021SpanbasedSP}, which introduced a method that learned to predict span trees over the input question that could be composed together to build complete logical forms. Their method achieves strong performance on the Geoquery SQL dataset \cite{finegan2018improving}; however, similar to \cite{rubin2021smbop}, their approach differs from ours in that they assume a one-to-one correspondence between disjoint spans of the question and disjoint subtrees of the logical form. Further, their method can only handle inputs whose parse falls into a restricted class of non-projective trees, where the authors note that extending their method to all classes of non-projective trees would prohibitively increase the time complexity of their parser. Such assumptions make it unclear how it would directly apply to a more complex dataset like CFQ.

There are several works targeted specifically towards compositional generalization \cite{gai2021grounded,liu2021learning,jambor2021lagr,guo2020hierarchical,herzig2021unlocking}. Though these systems have proven to be quite effective on compositional generalization datasets, they make significant task-specific assumptions that limit how they might be applied more broadly.

% GO THROUGH DIFFERENCES BETWEEN THIS AND \cite{zhou2021amr,zhou2021structure}

\section{Limitations}

The main limitation to our work lies in the handling of unordered $n$-ary relations. We hypothesize that the bottom-up paradigm performs best when there is one unambiguous logical form to generate for a particular question. While this is quite often true for semantic parsing, in our experience, unordered $n$-ary relations (e.g., \texttt{WHERE}) can quickly cause this not to be the case. With such relations, there tends to be a large number of correct logical forms for a particular question. In these situations, having so many candidate logical forms can cause significant issues in terms of runtime. A second limitation of our work is that it assumes the logical form will be given as a graph. Thus, there is an annotation burden on the users of this system that would not be present in systems that treat semantic parsing as a text-to-text problem (e.g., fine-tuned large language models).
\section{Conclusions}
\label{conclusions}

In this work, we introduced a novel, bottom-up decoding scheme that semi-autoregressively constructs logical forms in a way that facilitates compositional generalization. Key to our method was a lazy generation process designed to address the computational efficiency issues inherent to bottom-up parsing. We demonstrated our approach on five different datasets covering three logical formalisms and found that our approach was strongly competitive with neural models tailored to each task.

% Entries for the entire Anthology, followed by custom entries
\bibliography{anthology,main}

\begin{thebibliography}{50}
\expandafter\ifx\csname natexlab\endcsname\relax\def\natexlab#1{#1}\fi

\bibitem[{Alvarez-Melis and Jaakkola(2017)}]{AlvarezMelis2017TreestructuredDW}
David Alvarez-Melis and T.~Jaakkola. 2017.
\newblock Tree-structured decoding with doubly-recurrent neural networks.
\newblock In \emph{ICLR}.

\bibitem[{Andreas(2020)}]{andreas2020good}
Jacob Andreas. 2020.
\newblock Good-enough compositional data augmentation.
\newblock In \emph{Proceedings of the 58th Annual Meeting of the Association
  for Computational Linguistics}, pages 7556--7566.

\bibitem[{Bergen et~al.(2021)Bergen, O'Donnell, and
  Bahdanau}]{bergen2021systematic}
Leon Bergen, Timothy O'Donnell, and Dzmitry Bahdanau. 2021.
\newblock Systematic generalization with edge transformers.
\newblock \emph{Advances in Neural Information Processing Systems}, 34.

\bibitem[{Bogin et~al.(2022)Bogin, Gupta, and
  Berant}]{bogin-etal-2022-unobserved}
Ben Bogin, Shivanshu Gupta, and Jonathan Berant. 2022.
\newblock \href {https://aclanthology.org/2022.emnlp-main.175} {Unobserved
  local structures make compositional generalization hard}.
\newblock In \emph{Proceedings of the 2022 Conference on Empirical Methods in
  Natural Language Processing}, pages 2731--2747, Abu Dhabi, United Arab
  Emirates. Association for Computational Linguistics.

\bibitem[{Carpenter(1997)}]{Carpenter1997TypeLogicalS}
Bob Carpenter. 1997.
\newblock Type-logical semantics.

\bibitem[{Crouse(2021)}]{crouse2021question}
Maxwell Crouse. 2021.
\newblock \emph{Question-Answering with Structural Analogy}.
\newblock Ph.D. thesis, Northwestern University.

\bibitem[{Dahl et~al.(1994)Dahl, Bates, Brown, Fisher, Hunicke-Smith, Pallett,
  Pao, Rudnicky, and Shriberg}]{dahl1994expanding}
Deborah~A Dahl, Madeleine Bates, Michael~K Brown, William~M Fisher, Kate
  Hunicke-Smith, David~S Pallett, Christine Pao, Alexander Rudnicky, and
  Elizabeth Shriberg. 1994.
\newblock Expanding the scope of the atis task: The atis-3 corpus.
\newblock In \emph{Human Language Technology: Proceedings of a Workshop held at
  Plainsboro, New Jersey, March 8-11, 1994}.

\bibitem[{Dong and Lapata(2016)}]{dong-lapata-2016-language}
Li~Dong and Mirella Lapata. 2016.
\newblock \href {https://doi.org/10.18653/v1/P16-1004} {Language to logical
  form with neural attention}.
\newblock In \emph{Proceedings of the 54th Annual Meeting of the Association
  for Computational Linguistics (Volume 1: Long Papers)}, pages 33--43, Berlin,
  Germany. Association for Computational Linguistics.

\bibitem[{Dong and Lapata(2018)}]{dong-lapata-2018-coarse}
Li~Dong and Mirella Lapata. 2018.
\newblock \href {https://doi.org/10.18653/v1/P18-1068} {Coarse-to-fine decoding
  for neural semantic parsing}.
\newblock In \emph{Proceedings of the 56th Annual Meeting of the Association
  for Computational Linguistics (Volume 1: Long Papers)}, pages 731--742,
  Melbourne, Australia. Association for Computational Linguistics.

\bibitem[{Drozdov et~al.(2022)Drozdov, Sch{\"a}rli, Aky{\"u}rek, Scales, Song,
  Chen, Bousquet, and Zhou}]{drozdov2022compositional}
Andrew Drozdov, Nathanael Sch{\"a}rli, Ekin Aky{\"u}rek, Nathan Scales, Xinying
  Song, Xinyun Chen, Olivier Bousquet, and Denny Zhou. 2022.
\newblock Compositional semantic parsing with large language models.
\newblock \emph{arXiv preprint arXiv:2209.15003}.

\bibitem[{Finegan-Dollak et~al.(2018)Finegan-Dollak, Kummerfeld, Zhang,
  Ramanathan, Sadasivam, Zhang, and Radev}]{finegan2018improving}
Catherine Finegan-Dollak, Jonathan~K Kummerfeld, Li~Zhang, Karthik Ramanathan,
  Sesh Sadasivam, Rui Zhang, and Dragomir Radev. 2018.
\newblock Improving text-to-sql evaluation methodology.
\newblock In \emph{Proceedings of the 56th Annual Meeting of the Association
  for Computational Linguistics (Volume 1: Long Papers)}, pages 351--360.

\bibitem[{Furrer et~al.(2020)Furrer, van Zee, Scales, and
  Sch{\"a}rli}]{furrer2020compositional}
Daniel Furrer, Marc van Zee, Nathan Scales, and Nathanael Sch{\"a}rli. 2020.
\newblock Compositional generalization in semantic parsing: Pre-training vs.
  specialized architectures.
\newblock \emph{arXiv preprint arXiv:2007.08970}.

\bibitem[{Gai et~al.(2021)Gai, Jain, Zhang, Gonzalez, Song, and
  Stoica}]{gai2021grounded}
Yu~Gai, Paras Jain, Wendi Zhang, Joseph Gonzalez, Dawn Song, and Ion Stoica.
  2021.
\newblock Grounded graph decoding improves compositional generalization in
  question answering.
\newblock In \emph{Findings of the Association for Computational Linguistics:
  EMNLP 2021}, pages 1829--1838.

\bibitem[{Guo et~al.(2020)Guo, Lin, Lou, and Zhang}]{guo2020hierarchical}
Yinuo Guo, Zeqi Lin, Jian-Guang Lou, and Dongmei Zhang. 2020.
\newblock Hierarchical poset decoding for compositional generalization in
  language.
\newblock \emph{Advances in Neural Information Processing Systems}, 33.

\bibitem[{Herzig and Berant(2021)}]{Herzig2021SpanbasedSP}
Jonathan Herzig and Jonathan Berant. 2021.
\newblock Span-based semantic parsing for compositional generalization.
\newblock In \emph{ACL/IJCNLP}.

\bibitem[{Herzig et~al.(2021)Herzig, Shaw, Chang, Guu, Pasupat, and
  Zhang}]{herzig2021unlocking}
Jonathan Herzig, Peter Shaw, Ming-Wei Chang, Kelvin Guu, Panupong Pasupat, and
  Yuan Zhang. 2021.
\newblock Unlocking compositional generalization in pre-trained models using
  intermediate representations.
\newblock \emph{arXiv preprint arXiv:2104.07478}.

\bibitem[{Higgins et~al.(2018)Higgins, Sonnerat, Matthey, Pal, Burgess,
  Bo{\v{s}}njak, Shanahan, Botvinick, Hassabis, and Lerchner}]{higgins2018scan}
Irina Higgins, Nicolas Sonnerat, Loic Matthey, Arka Pal, Christopher~P Burgess,
  Matko Bo{\v{s}}njak, Murray Shanahan, Matthew Botvinick, Demis Hassabis, and
  Alexander Lerchner. 2018.
\newblock Scan: Learning hierarchical compositional visual concepts.
\newblock In \emph{International Conference on Learning Representations}.

\bibitem[{Hochreiter and Schmidhuber(1997)}]{hochreiter1997long}
Sepp Hochreiter and J{\"u}rgen Schmidhuber. 1997.
\newblock Long short-term memory.
\newblock \emph{Neural computation}, 9(8):1735--1780.

\bibitem[{Iyer et~al.(2017)Iyer, Konstas, Cheung, Krishnamurthy, and
  Zettlemoyer}]{iyer2017learning}
Srinivasan Iyer, Ioannis Konstas, Alvin Cheung, Jayant Krishnamurthy, and Luke
  Zettlemoyer. 2017.
\newblock Learning a neural semantic parser from user feedback.
\newblock In \emph{55th Annual Meeting of the Association for Computational
  Linguistics}.

\bibitem[{Jambor and Bahdanau(2021)}]{jambor2021lagr}
Dora Jambor and Dzmitry Bahdanau. 2021.
\newblock Lagr: Labeling aligned graphs for improving systematic generalization
  in semantic parsing.
\newblock \emph{arXiv preprint arXiv:2110.07572}.

\bibitem[{Jia and Liang(2016)}]{jia-liang-2016-data}
Robin Jia and Percy Liang. 2016.
\newblock \href {https://doi.org/10.18653/v1/P16-1002} {Data recombination for
  neural semantic parsing}.
\newblock In \emph{Proceedings of the 54th Annual Meeting of the Association
  for Computational Linguistics (Volume 1: Long Papers)}, pages 12--22, Berlin,
  Germany. Association for Computational Linguistics.

\bibitem[{Keysers et~al.(2019)Keysers, Sch{\"a}rli, Scales, Buisman, Furrer,
  Kashubin, Momchev, Sinopalnikov, Stafiniak, Tihon
  et~al.}]{keysers2019measuring}
Daniel Keysers, Nathanael Sch{\"a}rli, Nathan Scales, Hylke Buisman, Daniel
  Furrer, Sergii Kashubin, Nikola Momchev, Danila Sinopalnikov, Lukasz
  Stafiniak, Tibor Tihon, et~al. 2019.
\newblock Measuring compositional generalization: A comprehensive method on
  realistic data.
\newblock In \emph{International Conference on Learning Representations}.

\bibitem[{Kim and Linzen(2020)}]{kim2020cogs}
Najoung Kim and Tal Linzen. 2020.
\newblock Cogs: A compositional generalization challenge based on semantic
  interpretation.
\newblock \emph{arXiv preprint arXiv:2010.05465}.

\bibitem[{Krishnamurthy et~al.(2017)Krishnamurthy, Dasigi, and
  Gardner}]{krishnamurthy-etal-2017-neural}
Jayant Krishnamurthy, Pradeep Dasigi, and Matt Gardner. 2017.
\newblock \href {https://doi.org/10.18653/v1/D17-1160} {Neural semantic parsing
  with type constraints for semi-structured tables}.
\newblock In \emph{Proceedings of the 2017 Conference on Empirical Methods in
  Natural Language Processing}, pages 1516--1526, Copenhagen, Denmark.
  Association for Computational Linguistics.

\bibitem[{Kwiatkowski et~al.(2011)Kwiatkowski, Zettlemoyer, Goldwater, and
  Steedman}]{kwiatkowski2011lexical}
Tom Kwiatkowski, Luke Zettlemoyer, Sharon Goldwater, and Mark Steedman. 2011.
\newblock Lexical generalization in ccg grammar induction for semantic parsing.
\newblock In \emph{Proceedings of the 2011 Conference on Empirical Methods in
  Natural Language Processing}, pages 1512--1523.

\bibitem[{Lake and Baroni(2018)}]{lake2018generalization}
Brenden Lake and Marco Baroni. 2018.
\newblock Generalization without systematicity: On the compositional skills of
  sequence-to-sequence recurrent networks.
\newblock In \emph{International conference on machine learning}, pages
  2873--2882. PMLR.

\bibitem[{Lake et~al.(2017)Lake, Ullman, Tenenbaum, and
  Gershman}]{lake2017building}
Brenden~M Lake, Tomer~D Ullman, Joshua~B Tenenbaum, and Samuel~J Gershman.
  2017.
\newblock Building machines that learn and think like people.
\newblock \emph{Behavioral and brain sciences}, 40.

\bibitem[{Li et~al.(2020)Li, Wu, Feng, Xu, Xu, and Zhong}]{li2020graph}
Shucheng Li, Lingfei Wu, Shiwei Feng, Fangli Xu, Fengyuan Xu, and Sheng Zhong.
  2020.
\newblock Graph-to-tree neural networks for learning structured input-output
  translation with applications to semantic parsing and math word problem.
\newblock \emph{EMNLP}.

\bibitem[{Liang et~al.(2013)Liang, Jordan, and Klein}]{liang2013learning}
Percy Liang, Michael~I Jordan, and Dan Klein. 2013.
\newblock Learning dependency-based compositional semantics.
\newblock \emph{Computational Linguistics}, 39(2):389--446.

\bibitem[{Liu et~al.(2021)Liu, An, Lin, Liu, Chen, Lou, Wen, Zheng, and
  Zhang}]{liu2021learning}
Chenyao Liu, Shengnan An, Zeqi Lin, Qian Liu, Bei Chen, Jian-Guang Lou, Lijie
  Wen, Nanning Zheng, and Dongmei Zhang. 2021.
\newblock Learning algebraic recombination for compositional generalization.
\newblock In \emph{Findings of the Association for Computational Linguistics:
  ACL-IJCNLP 2021}, pages 1129--1144.

\bibitem[{McCarthy(1978)}]{mccarthy1978history}
John McCarthy. 1978.
\newblock History of lisp.
\newblock In \emph{History of programming languages}, pages 173--185.

\bibitem[{McCarthy(1983)}]{mccarthy1983recursive}
John McCarthy. 1983.
\newblock Recursive functions of symbolic expressions.
\newblock In \emph{Programming Languages}, pages 175--186. Springer.

\bibitem[{Montague(1973)}]{montague1973proper}
Richard Montague. 1973.
\newblock The proper treatment of quantification in ordinary english.
\newblock In \emph{Approaches to natural language}, pages 221--242. Springer.

\bibitem[{Oren et~al.(2020)Oren, Herzig, Gupta, Gardner, and
  Berant}]{oren2020improving}
Inbar Oren, Jonathan Herzig, Nitish Gupta, Matt Gardner, and Jonathan Berant.
  2020.
\newblock Improving compositional generalization in semantic parsing.
\newblock In \emph{EMNLP (Findings)}.

\bibitem[{Qiu et~al.(2022)Qiu, Shaw, Pasupat, Shi, Herzig, Pitler, Sha, and
  Toutanova}]{qiu-etal-2022-evaluating}
Linlu Qiu, Peter Shaw, Panupong Pasupat, Tianze Shi, Jonathan Herzig, Emily
  Pitler, Fei Sha, and Kristina Toutanova. 2022.
\newblock \href {https://aclanthology.org/2022.emnlp-main.624} {Evaluating the
  impact of model scale for compositional generalization in semantic parsing}.
\newblock In \emph{Proceedings of the 2022 Conference on Empirical Methods in
  Natural Language Processing}, pages 9157--9179, Abu Dhabi, United Arab
  Emirates. Association for Computational Linguistics.

\bibitem[{Raffel et~al.(2019)Raffel, Shazeer, Roberts, Lee, Narang, Matena,
  Zhou, Li, and Liu}]{raffel2019exploring}
Colin Raffel, Noam Shazeer, Adam Roberts, Katherine Lee, Sharan Narang, Michael
  Matena, Yanqi Zhou, Wei Li, and Peter~J Liu. 2019.
\newblock Exploring the limits of transfer learning with a unified text-to-text
  transformer.
\newblock \emph{arXiv preprint arXiv:1910.10683}.

\bibitem[{Rubin and Berant(2021)}]{rubin2021smbop}
Ohad Rubin and Jonathan Berant. 2021.
\newblock Smbop: Semi-autoregressive bottom-up semantic parsing.
\newblock In \emph{Proceedings of the 5th Workshop on Structured Prediction for
  NLP (SPNLP 2021)}, pages 12--21.

\bibitem[{Shaw et~al.(2021)Shaw, Chang, Pasupat, and
  Toutanova}]{shaw2021compositional}
Peter Shaw, Ming-Wei Chang, Panupong Pasupat, and Kristina Toutanova. 2021.
\newblock Compositional generalization and natural language variation: Can a
  semantic parsing approach handle both?
\newblock In \emph{Proceedings of the 59th Annual Meeting of the Association
  for Computational Linguistics and the 11th International Joint Conference on
  Natural Language Processing (Volume 1: Long Papers)}, pages 922--938.

\bibitem[{Shaw et~al.(2018)Shaw, Uszkoreit, and Vaswani}]{shaw2018self}
Peter Shaw, Jakob Uszkoreit, and Ashish Vaswani. 2018.
\newblock Self-attention with relative position representations.
\newblock In \emph{Proceedings of the 2018 Conference of the North American
  Chapter of the Association for Computational Linguistics: Human Language
  Technologies, Volume 2 (Short Papers)}, pages 464--468.

\bibitem[{Vaswani et~al.(2017)Vaswani, Shazeer, Parmar, Uszkoreit, Jones,
  Gomez, Kaiser, and Polosukhin}]{vaswani2017attention}
Ashish Vaswani, Noam Shazeer, Niki Parmar, Jakob Uszkoreit, Llion Jones,
  Aidan~N Gomez, {\L}ukasz Kaiser, and Illia Polosukhin. 2017.
\newblock Attention is all you need.
\newblock In \emph{Advances in neural information processing systems}, pages
  5998--6008.

\bibitem[{Wong and Mooney(2007)}]{wong2007learning}
Yuk~Wah Wong and Raymond Mooney. 2007.
\newblock Learning synchronous grammars for semantic parsing with lambda
  calculus.
\newblock In \emph{Proceedings of the 45th Annual Meeting of the Association of
  Computational Linguistics}, pages 960--967.

\bibitem[{Xiao et~al.(2016)Xiao, Dymetman, and
  Gardent}]{xiao-etal-2016-sequence}
Chunyang Xiao, Marc Dymetman, and Claire Gardent. 2016.
\newblock \href {https://doi.org/10.18653/v1/P16-1127} {Sequence-based
  structured prediction for semantic parsing}.
\newblock In \emph{Proceedings of the 54th Annual Meeting of the Association
  for Computational Linguistics (Volume 1: Long Papers)}, pages 1341--1350,
  Berlin, Germany. Association for Computational Linguistics.

\bibitem[{Xu et~al.(2018)Xu, Wu, Wang, Feng, Witbrock, and
  Sheinin}]{xu2018graph2seq}
Kun Xu, Lingfei Wu, Zhiguo Wang, Yansong Feng, Michael Witbrock, and Vadim
  Sheinin. 2018.
\newblock Graph2seq: Graph to sequence learning with attention-based neural
  networks.
\newblock \emph{arXiv preprint arXiv:1804.00823}.

\bibitem[{Yu et~al.(2018)Yu, Zhang, Yang, Yasunaga, Wang, Li, Ma, Li, Yao,
  Roman, Zhang, and Radev}]{Yu2018SpiderAL}
Tao Yu, Rui Zhang, Kai-Chou Yang, Michihiro Yasunaga, Dongxu Wang, Zifan Li,
  James Ma, Irene~Z Li, Qingning Yao, Shanelle Roman, Zilin Zhang, and
  Dragomir~R. Radev. 2018.
\newblock Spider: A large-scale human-labeled dataset for complex and
  cross-domain semantic parsing and text-to-sql task.
\newblock In \emph{EMNLP}.

\bibitem[{Zelle and Mooney(1996)}]{zelle1996learning}
John~M Zelle and Raymond~J Mooney. 1996.
\newblock Learning to parse database queries using inductive logic programming.
\newblock In \emph{Proceedings of the national conference on artificial
  intelligence}, pages 1050--1055.

\bibitem[{Zettlemoyer and Collins(2007)}]{zettlemoyer2007online}
Luke Zettlemoyer and Michael Collins. 2007.
\newblock Online learning of relaxed ccg grammars for parsing to logical form.
\newblock In \emph{Proceedings of the 2007 Joint Conference on Empirical
  Methods in Natural Language Processing and Computational Natural Language
  Learning (EMNLP-CoNLL)}, pages 678--687.

\bibitem[{Zhao and Huang(2015)}]{zhao2015type}
Kai Zhao and Liang Huang. 2015.
\newblock Type-driven incremental semantic parsing with polymorphism.
\newblock In \emph{Proceedings of the 2015 Conference of the North American
  Chapter of the Association for Computational Linguistics: Human Language
  Technologies}, pages 1416--1421.

\bibitem[{Zheng and Lapata(2020)}]{zheng2020compositional}
Hao Zheng and Mirella Lapata. 2020.
\newblock Compositional generalization via semantic tagging.
\newblock \emph{arXiv preprint arXiv:2010.11818}.

\bibitem[{Zhou et~al.(2021{\natexlab{a}})Zhou, Naseem, Astudillo, and
  Florian}]{zhou2021amr}
Jiawei Zhou, Tahira Naseem, Ram{\'o}n~Fernandez Astudillo, and Radu Florian.
  2021{\natexlab{a}}.
\newblock Amr parsing with action-pointer transformer.
\newblock In \emph{Proceedings of the 2021 Conference of the North American
  Chapter of the Association for Computational Linguistics: Human Language
  Technologies}, pages 5585--5598.

\bibitem[{Zhou et~al.(2021{\natexlab{b}})Zhou, Naseem, Astudillo, Lee, Florian,
  and Roukos}]{zhou2021structure}
Jiawei Zhou, Tahira Naseem, Ram{\'o}n~Fernandez Astudillo, Young-Suk Lee, Radu
  Florian, and Salim Roukos. 2021{\natexlab{b}}.
\newblock Structure-aware fine-tuning of sequence-to-sequence transformers for
  transition-based amr parsing.
\newblock In \emph{Proceedings of the 2021 Conference on Empirical Methods in
  Natural Language Processing}, pages 6279--6290.

\end{thebibliography}
\bibliographystyle{acl_natbib}

\appendix

\newpage
\section{Appendix}

\subsection{Dataset Sizes}

\begin{table}[b]
\begin{center}
\begin{small}
\begin{tabular}{lccc}
\toprule
Dataset & Split & Train & Test \\
\midrule
CFQ & MCD1 & 95k & 12k \\
 & MCD2 & 95k & 12k \\ 
 & MCD3 & 95k & 12k \\
Text-To-SQL & ATIS & 4812 & 347 \\
 & Geoquery & 539 & 182 \\
 & Scholar & 408 & 315 \\
Geoquery & - & 600 & 280 \\
\bottomrule
\end{tabular}
\end{small}
\end{center}
\caption{Dataset sizes}
\label{tbl:hparams_lstm}
\end{table}

Geoquery \cite{zelle1996learning} is a standard benchmark dataset for semantic parsing that consists of 880 geography-related questions. We use the standard \cite{zettlemoyer2007online} split of 600 train questions and 280 test questions and the logical form representation of \cite{dong-lapata-2016-language,kwiatkowski2011lexical}, which is a variant of typed lambda calculus \cite{Carpenter1997TypeLogicalS}.

\subsection{Training Details}
All models were trained on an HPC cluster, where each training run was provided with 100 GB RAM, 2 CPUs, and 1 V100 GPU. The longest training run (on CFQ with T5-base) would complete within 24 hours.

\subsection{Model Sizes}

The T5-base models have around 220 million parameters. As our model shares all parameters with T5 except for the final attention-based pointer module, a rough estimate is that our model would have approximately 221 million parameters (with $768 \times 768$ additional parameters from the attention module).

\subsection{Hyperparameters}
\label{sec:hyperparams}

We did not do much in the way of hyperparameter tuning, instead opting to base most values off of the original CFQ \cite{keysers2019measuring} work and sharing most hyperparameter settings between our experiments. In Tables \ref{tbl:hparams_lstm} and \ref{tbl:hparams_t5}, we list the hyperparameters used in both experiments.

\begin{table*}[t]
\begin{center}
\begin{small}
\begin{tabular}{lccc}
\toprule
Hyperparameter & CFQ & Text-to-SQL & Geoquery \\
\midrule
Embedding dimensionality & 256 & 256 & 256 \\
Number of MHA heads & 8 & 8 & 8 \\
Dimensionality of FFN hidden layers & 1024 & 1024 & 1024 \\
Encoder Dropout & 0.4 & 0.4 & 0.4 \\
Decoder Dropout & 0.1 & 0.1 & 0.1 \\
Batch size & 32 & 32 & 32 \\
Learning rate & 0.0001 & 0.001 & 0.001 \\
Number of training epochs & 100 & 500 & 500 \\
\bottomrule
\end{tabular}
\end{small}
\end{center}
\caption{Hyperparameters for CFQ, Text-to-SQL, and Geoquery experiments when an LSTM encoder was used}
\label{tbl:hparams_lstm}
\end{table*}

\begin{table*}[t]
\begin{center}
\begin{small}
\begin{tabular}{lcc}
\toprule
Hyperparameter & CFQ & Text-to-SQL \\
\midrule
Embedding dimensionality & 768 & 768 \\
Number of MHA heads & 12 & 12 \\
Dimensionality of FFN hidden layers & 3072 & 3072 \\
Encoder Dropout & 0.1 & 0.1 \\
Decoder Dropout & 0.1 & 0.1 \\
Batch size & 16 & 16 \\
Learning rate & 0.0001 & 0.0001 \\
Number of training epochs & 20 & 500 \\
\bottomrule
\end{tabular}
\end{small}
\end{center}
\caption{Hyperparameters for CFQ and Text-to-SQL experiments when T5-base was used}
\label{tbl:hparams_t5}
\end{table*}

% \subsection{CFQ Dataset}

% \begin{figure}[t]
% \begin{subfigure}{\columnwidth}
% \begin{center}
% \begin{tcolorbox}[width=0.95\columnwidth,colback=white!90!black]
% % {\texttt{SELECT count ( * ) WHERE \{ \\ \hspace*{1pt} ?x0 producer.film M0 .\\ \hspace*{1pt} ?x0 writer.film M0 .\\ \hspace*{1pt} M1 film.directed\_by ?x0 .\\ \hspace*{1pt} M1 film.edited\_by ?x0 \\ \} }}
% {\texttt{SELECT count ( * ) WHERE \{ \\ \hspace*{1pt} M0 directed\_by M1 .\\ \hspace*{1pt} M0 directed\_by M2 .\\ \hspace*{1pt} M0 edited\_by M1 .\\ \hspace*{1pt} M0 edited\_by M2 .\\ \hspace*{1pt} M0 prod\_companies M1 .\\ \hspace*{1pt} M0 prod\_companies M2 \\ \}
% }} \\ \\
% {\texttt{(SELECT (count *) (WHERE \\ \hspace*{1pt} (directed\_by M0 M1) \\ \hspace*{1pt} (directed\_by M0 M2)\\ \hspace*{1pt} (edited\_by M0 M1)\\ \hspace*{1pt} (edited\_by M0 M2)\\ \hspace*{1pt} (prod\_companies M0 M1)\\ \hspace*{1pt} (prod\_companies M0 M2)\\ ))
% }}
% \end{tcolorbox}
% \end{center}
% \end{subfigure}

% \caption{SPARQL query and its corresponding s-expression for the CFQ question "Was M0 produced, directed, and edited by M1 and M2"}
% \label{fig:cfq_repr}
% \end{figure}

% Converting CFQ to s-expressions is straightforward, where the 

% \subsection{Text-to-SQL Datasets}

% sql.Identifier, sql.Statement, sql.Parenthesis, sql.Comparison

% \subsection{Geoquery Dataset}

%  In our experiments, this version of Geoquery serves to validate that our method can operate in small-data settings and with other logical formalisms than SPARQL and SQL.

\end{document}